\documentclass{article}

\usepackage{microtype}
\usepackage{graphicx}
\usepackage{subfigure}
\usepackage{booktabs} 

\usepackage{hyperref}
\usepackage{enumitem}
\usepackage{xcolor}

\usepackage[accepted]{icml2024}

\usepackage{amsmath}
\usepackage{amssymb}
\usepackage{mathtools}
\usepackage{amsthm}
\usepackage{ulem} 
\usepackage{multirow}

\usepackage[capitalize,noabbrev]{cleveref}

\theoremstyle{plain}
\newtheorem{theorem}{Theorem}[section]
\newtheorem{proposition}[theorem]{Proposition}

\newtheorem{corollary}[theorem]{Corollary}
\theoremstyle{definition}

\theoremstyle{remark}

\usepackage[textsize=tiny]{todonotes}

\icmltitlerunning{PDHG-Unrolled Learning-to-Optimize Method for Large-Scale Linear Programming}

\begin{document}
\global\long\def\inprod#1#2{\left\langle #1,#2\right\rangle }%
\global\long\def\inner#1#2{\langle#1,#2\rangle}%
\global\long\def\binner#1#2{\big\langle#1,#2\big\rangle}%
\global\long\def\norm#1{\Vert#1\Vert}%
\global\long\def\bnorm#1{\big\Vert#1\big\Vert}%
\global\long\def\Bnorm#1{\Big\Vert#1\Big\Vert}%
\global\long\def\red#1{\textcolor{red}{#1}}%
\global\long\def\blue#1{\textcolor{blue}{#1}}%

\global\long\def\brbra#1{\big(#1\big)}%
\global\long\def\Brbra#1{\Big(#1\Big)}%
\global\long\def\rbra#1{(#1)}%

\global\long\def\sbra#1{[#1]}%
\global\long\def\bsbra#1{\big[#1\big]}%
\global\long\def\Bsbra#1{\Big[#1\Big]}%
\global\long\def\abs#1{\vert#1\vert}%
\global\long\def\babs#1{\big\vert#1\big\vert}%

\global\long\def\cbra#1{\{#1\}}%
\global\long\def\bcbra#1{\big\{#1\big\}}%
\global\long\def\Bcbra#1{\Big\{#1\Big\}}%
\global\long\def\vertiii#1{\left\vert \kern-0.25ex  \left\vert \kern-0.25ex  \left\vert #1\right\vert \kern-0.25ex  \right\vert \kern-0.25ex  \right\vert }%
\global\long\def\matr#1{\bm{#1}}%
\global\long\def\til#1{\tilde{#1}}%
\global\long\def\wtil#1{\widetilde{#1}}%
\global\long\def\wh#1{\widehat{#1}}%
\global\long\def\mcal#1{\mathcal{#1}}%
\global\long\def\mbb#1{\mathbb{#1}}%
\global\long\def\mtt#1{\mathtt{#1}}%
\global\long\def\ttt#1{\texttt{#1}}%
\global\long\def\dtxt{\textrm{d}}%
\global\long\def\bignorm#1{\bigl\Vert#1\bigr\Vert}%
\global\long\def\Bignorm#1{\Bigl\Vert#1\Bigr\Vert}%
\global\long\def\rmn#1#2{\mathbb{R}^{#1\times#2}}%
\global\long\def\deri#1#2{\frac{d#1}{d#2}}%
\global\long\def\pderi#1#2{\frac{\partial#1}{\partial#2}}%
\global\long\def\limk{\lim_{k\rightarrow\infty}}%
\global\long\def\trans{\textrm{T}}%
\global\long\def\onebf{\mathbf{1}}%
\global\long\def\Bbb{\mathbb{B}}%
\global\long\def\hbf{\mathbf{h}}%
\global\long\def\zerobf{\mathbf{0}}%
\global\long\def\zero{\bm{0}}%

\global\long\def\Euc{\mathrm{E}}%
\global\long\def\Expe{\mathbb{E}}%
\global\long\def\rank{\mathrm{rank}}%
\global\long\def\range{\mathrm{range}}%
\global\long\def\diam{\mathrm{diam}}%
\global\long\def\epi{\mathrm{epi} }%
\global\long\def\inte{\operatornamewithlimits{int}}%
\global\long\def\dist{\operatornamewithlimits{dist}}%
\global\long\def\proj{\operatorname{Proj}}%
\global\long\def\cov{\mathrm{Cov}}%
\global\long\def\argmin{\operatornamewithlimits{argmin}}%
\global\long\def\argmax{\operatornamewithlimits{argmax}}%
\global\long\def\tr{\operatornamewithlimits{tr}}%
\global\long\def\dis{\operatornamewithlimits{dist}}%
\global\long\def\sign{\operatornamewithlimits{sign}}%
\global\long\def\prob{\mathrm{Prob}}%
\global\long\def\st{\operatornamewithlimits{s.t.}}%
\global\long\def\dom{\mathrm{dom}}%
\global\long\def\prox{\mathrm{prox}}%
\global\long\def\diag{\mathrm{diag}}%
\global\long\def\and{\mathrm{and}}%
\global\long\def\aleq{\overset{(a)}{\leq}}%
\global\long\def\aeq{\overset{(a)}{=}}%
\global\long\def\ageq{\overset{(a)}{\geq}}%
\global\long\def\bleq{\overset{(b)}{\leq}}%
\global\long\def\beq{\overset{(b)}{=}}%
\global\long\def\bgeq{\overset{(b)}{\geq}}%
\global\long\def\cleq{\overset{(c)}{\leq}}%
\global\long\def\ceq{\overset{(c)}{=}}%
\global\long\def\cgeq{\overset{(c)}{\geq}}%
\global\long\def\dleq{\overset{(d)}{\leq}}%
\global\long\def\deq{\overset{(d)}{=}}%
\global\long\def\dgeq{\overset{(d)}{\geq}}%
\global\long\def\eleq{\overset{(e)}{\leq}}%
\global\long\def\eeq{\overset{(e)}{=}}%
\global\long\def\egeq{\overset{(e)}{\geq}}%
\global\long\def\fleq{\overset{(f)}{\leq}}%
\global\long\def\feq{\overset{(f)}{=}}%
\global\long\def\fgeq{\overset{(f)}{\geq}}%
\global\long\def\gleq{\overset{(g)}{\leq}}%
\global\long\def\as{\textup{a.s.}}%
\global\long\def\ae{\textup{a.e.}}%
\global\long\def\Var{\operatornamewithlimits{Var}}%
\global\long\def\clip{\operatorname{clip}}%
\global\long\def\conv{\operatorname{conv}}%
\global\long\def\Cov{\operatornamewithlimits{Cov}}%
\global\long\def\raw{\rightarrow}%
\global\long\def\law{\leftarrow}%
\global\long\def\Raw{\Rightarrow}%
\global\long\def\Law{\Leftarrow}%
\global\long\def\vep{\varepsilon}%
\global\long\def\dom{\operatornamewithlimits{dom}}%
\global\long\def\tsum{{\textstyle {\sum}}}%
\global\long\def\Cbb{\mathbb{C}}%
\global\long\def\Ebb{\mathbb{E}}%
\global\long\def\Fbb{\mathbb{F}}%
\global\long\def\Nbb{\mathbb{N}}%
\global\long\def\Rbb{\mathbb{R}}%
\global\long\def\extR{\widebar{\mathbb{R}}}%
\global\long\def\Pbb{\mathbb{P}}%
\global\long\def\Zbb{\mathbb{Z}}%
\global\long\def\Mrm{\mathrm{M}}%
\global\long\def\Acal{\mathcal{A}}%
\global\long\def\Bcal{\mathcal{B}}%
\global\long\def\Ccal{\mathcal{C}}%
\global\long\def\Dcal{\mathcal{D}}%
\global\long\def\Ecal{\mathcal{E}}%
\global\long\def\Fcal{\mathcal{F}}%
\global\long\def\Gcal{\mathcal{G}}%
\global\long\def\Hcal{\mathcal{H}}%
\global\long\def\Ical{\mathcal{I}}%
\global\long\def\Kcal{\mathcal{K}}%
\global\long\def\Lcal{\mathcal{L}}%
\global\long\def\Mcal{\mathcal{M}}%
\global\long\def\Ncal{\mathcal{N}}%
\global\long\def\Ocal{\mathcal{O}}%
\global\long\def\Pcal{\mathcal{P}}%
\global\long\def\Scal{\mathcal{S}}%
\global\long\def\Tcal{\mathcal{T}}%
\global\long\def\Xcal{\mathcal{X}}%
\global\long\def\Ycal{\mathcal{Y}}%
\global\long\def\Zcal{\mathcal{Z}}%
\global\long\def\i{i}%

\global\long\def\abf{\mathbf{a}}%
\global\long\def\bbf{\mathbf{b}}%
\global\long\def\cbf{\mathbf{c}}%
\global\long\def\fbf{\mathbf{f}}%
\global\long\def\qbf{\mathbf{q}}%
\global\long\def\gbf{\mathbf{g}}%
\global\long\def\ebf{\mathbf{e}}%
\global\long\def\lambf{\bm{\lambda}}%
\global\long\def\alphabf{\bm{\alpha}}%
\global\long\def\sigmabf{\bm{\sigma}}%
\global\long\def\thetabf{\bm{\theta}}%
\global\long\def\deltabf{\bm{\delta}}%
\global\long\def\lbf{\mathbf{l}}%
\global\long\def\ubf{\mathbf{u}}%
\global\long\def\pbf{\mathbf{\mathbf{p}}}%
\global\long\def\vbf{\mathbf{v}}%
\global\long\def\wbf{\mathbf{w}}%
\global\long\def\xbf{\mathbf{x}}%
\global\long\def\ybf{\mathbf{y}}%
\global\long\def\zbf{\mathbf{z}}%
\global\long\def\dbf{\mathbf{d}}%
\global\long\def\Wbf{\mathbf{W}}%
\global\long\def\Abf{\mathbf{A}}%
\global\long\def\Ubf{\mathbf{U}}%
\global\long\def\Pbf{\mathbf{P}}%
\global\long\def\Ibf{\mathbf{I}}%
\global\long\def\Ebf{\mathbf{E}}%
\global\long\def\sbf{\mathbf{s}}%
\global\long\def\Mbf{\mathbf{M}}%
\global\long\def\Nbf{\mathbf{N}}%
\global\long\def\Dbf{\mathbf{D}}%
\global\long\def\Qbf{\mathbf{Q}}%
\global\long\def\Lbf{\mathbf{L}}%
\global\long\def\Pbf{\mathbf{P}}%
\global\long\def\Xbf{\mathbf{X}}%
\global\long\def\Bbf{\mathbf{B}}%
\global\long\def\zerobf{\mathbf{0}}%
\global\long\def\onebf{\mathbf{1}}%


\global\long\def\abm{\bm{a}}%
\global\long\def\bbm{\bm{b}}%
\global\long\def\cbm{\bm{c}}%
\global\long\def\dbm{\bm{d}}%
\global\long\def\ebm{\bm{e}}%
\global\long\def\fbm{\bm{f}}%
\global\long\def\gbm{\bm{g}}%
\global\long\def\hbm{\bm{h}}%
\global\long\def\pbm{\bm{p}}%
\global\long\def\qbm{\bm{q}}%
\global\long\def\rbm{\bm{r}}%
\global\long\def\sbm{\bm{s}}%
\global\long\def\tbm{\bm{t}}%
\global\long\def\ubm{\bm{u}}%
\global\long\def\vbm{\bm{v}}%
\global\long\def\wbm{\bm{w}}%
\global\long\def\xbm{\bm{x}}%
\global\long\def\ybm{\bm{y}}%
\global\long\def\zbm{\bm{z}}%
\global\long\def\Abm{\bm{A}}%
\global\long\def\Bbm{\bm{B}}%
\global\long\def\Cbm{\bm{C}}%
\global\long\def\Dbm{\bm{D}}%
\global\long\def\Ebm{\bm{E}}%
\global\long\def\Fbm{\bm{F}}%
\global\long\def\Gbm{\bm{G}}%
\global\long\def\Hbm{\bm{H}}%
\global\long\def\Ibm{\bm{I}}%
\global\long\def\Jbm{\bm{J}}%
\global\long\def\Lbm{\bm{L}}%
\global\long\def\Obm{\bm{O}}%
\global\long\def\Pbm{\bm{P}}%
\global\long\def\Qbm{\bm{Q}}%
\global\long\def\Rbm{\bm{R}}%
\global\long\def\Ubm{\bm{U}}%
\global\long\def\Vbm{\bm{V}}%
\global\long\def\Wbm{\bm{W}}%
\global\long\def\Xbm{\bm{X}}%
\global\long\def\Ybm{\bm{Y}}%
\global\long\def\Zbm{\bm{Z}}%
\global\long\def\lambm{\bm{\lambda}}%
\global\long\def\alphabm{\bm{\alpha}}%
\global\long\def\albm{\bm{\alpha}}%
\global\long\def\taubm{\bm{\tau}}%
\global\long\def\mubm{\bm{\mu}}%
\global\long\def\inftybm{\bm{\infty}}%
\global\long\def\yrm{\mathrm{y}}%

\twocolumn[
\icmltitle{PDHG-Unrolled Learning-to-Optimize Method \\
    for Large-Scale Linear Programming}



\icmlsetsymbol{equal}{*}

\begin{icmlauthorlist}
\icmlauthor{Bingheng Li}{equal,msu}
\icmlauthor{Linxin Yang}{equal,sds,sribd}
\icmlauthor{Yupeng Chen}{sribd}
\icmlauthor{Senmiao Wang}{sds}
\icmlauthor{Haitao Mao}{msu}
\icmlauthor{Qian Chen}{sribd,sse}
\icmlauthor{Yao Ma}{rpi}
\icmlauthor{Akang Wang}{sds,sribd}
\icmlauthor{Tian Ding}{sribd}
\icmlauthor{Jiliang Tang}{msu}
\icmlauthor{Ruoyu Sun}{sds,sribd}

\end{icmlauthorlist}

\icmlaffiliation{msu}{Michigan State University}
\icmlaffiliation{sds}{School of Data Science, The Chinese University of Hong Kong, Shenzhen, China }
\icmlaffiliation{sse}{School of Science and Engineering, The Chinese University of Hong Kong, Shenzhen, China }
\icmlaffiliation{sribd}{Shenzhen International Center For Industrial and Applied Mathematics, Shenzhen Research Institute of Big Data}
\icmlaffiliation{rpi}{Rensselaer Polytechnic Institute}

\icmlcorrespondingauthor{Ruoyu Sun}{sunruoyu@cuhk.edu.cn}

\icmlkeywords{Machine Learning, ICML}

\vskip 0.3in
]



\printAffiliationsAndNotice{\icmlEqualContribution} 

\begin{abstract}

Solving large-scale linear programming (LP) problems is an important task in various areas such as communication networks, power systems, finance, and logistics.
 Recently, two distinct approaches have emerged to expedite 
 LP solving: (i) First-order methods (FOMs); (ii) Learning to optimize (L2O). 
In this work, we propose a FOM-unrolled neural network (NN) 
called PDHG-Net, and propose a two-stage L2O method to solve
large-scale LP problems. 
The new architecture PDHG-Net is designed by unrolling the recently emerged PDHG method into a neural network, combined with channel-expansion techniques borrowed from graph neural networks.
We prove that the proposed PDHG-Net can recover
PDHG algorithm, thus can approximate optimal solutions of LP instances
with a polynomial number of neurons. 
We propose a two-stage inference approach: first use PDHG-Net to generate an approximate solution, and then apply the PDHG algorithm to further improve the solution.
Experiments show that our approach can significantly accelerate LP solving, achieving up to a 3$\times$ speedup compared to FOMs for large-scale LP problems. 

\end{abstract}
\section{Introduction}

Linear programming (LP)~\cite{luenberger1984linear,dantzig1998linear,schrijver1998theory} has wide applications in various areas such as 
commnication systems, power systems, logistics and finance~\cite{charnes1957management,rabinowitz1968applications,garver1970transmission,dorfman1987linear}. 
In the era of big data, the size of LP problems has increased
dramatically, leading to a growing interest in accelerating and scaling up LP algorithms ~\cite{todd1983large,bixby1992very,basu2020eclipse,applegate2021practical,deng2022new,hinder2023worst}.
Classical algorithms such as the interior-point methods do not scale up
due to the expensive per-iteration cost of solving linear systems.
Recently, first-order methods (FOMs) such as Primal-Dual Hybrid Gradient (PDHG) have shown considerable potential in solving large-scale LP problems~\cite{chambolle2016ergodic,applegate2021practical,fercoq2021quadratic,lu2022infimal,applegate2023faster,lu2023cupdlp,lu2023practical}.


Alongside first-order methods, 
``learning to optimize'' (L2O) paradigm emerges as a promising approach to speed up optimization algorithms \cite{nair2020solving, bengio2021machine, chen2022learning}.
L2O is an orthogonal line of pursuit compared to 
FOMs: L2O utilizes the information of existing
 problems to help solve new problems, 
 thus can potentially be applied to almost any class of optimization
 problems. 
 Interestingly, in the development of L2O methods, most efforts
 have been spent on integer programming, combinatorial optimization, and non-smooth optimization~\cite{gupta2020hybrid,gasse2022machine,liu2019alista}
but rarely on the more classical problem class LP problems. 
 Only until very recently, there have been some explorations
 on L2O for LP
~\cite{li2022learning,chen2022representing,kuang2023accelerate,fan2023smart,liu2023learning,qian2023exploring}.
 


We notice that L2O-for-LP works have primarily focused on the emulation of interior-point methods~\cite{qian2023exploring} and simplex methods~\cite{fan2023smart,liu2023learning}.
Motivated by recent progress on FOMs for LP, 
we ask the following question:  
\begin{equation*}
\begin{aligned}
& \textit{Can we design L2O methods to emulate FOMs}\\ 
& \quad\quad \ \textit{to solve large-scale LP problems?} 
\end{aligned} 
\end{equation*}

In this paper, we introduce a FOM-unrolled neural network architecture
PDHG-Net and a two-stage L2O framework for solving large-scale LP programs. 
To design the architecture, we unroll the PDHG algorithm into a neural network (NN) by turning the primal and dual updates into NN blocks,
and also incorporate the channel-expansion techniques from graph neural networks. 
The training part is similar to the standard L2O framework as we use a large number of LP instances to train PDHG-Net.
The inference part consists of two stages, which is
different from the standard L2O framework.
In the first stage, we use PDHG-Net to generate an approximately
optimal solution.
In the second stage, to enhance the solution precision, we refine the solution of PDHG-Net through PDLP. 
Our inference stage is a hybrid method of neural net 
and PDLP.

Our main contributions are summarized as follows:

\begin{itemize}[leftmargin=*]
    \item  \textbf{PDHG-Net Architecture: }  We propose PDHG-Net, a new neural network architecture inspired by the PDHG algorithm. In PDHG-Net, we replace projection functions with ReLU activations and incorporate channel expansion.
    
    
    \vspace{-1mm}
    \item \textbf{Theoretical Foundation: } Our theoretical analysis establishes an exact alignment between PDHG-Net and the standard PDHG algorithm. 
    Furthermore, given an approximation error threshold $\epsilon$, we theoretically prove that a PDHG-Net with $\mathcal{O}(1/\epsilon)$ number of neurons can approximate the optimal solution of LP problems to $\epsilon$-accuracy. 
    
    \item \textbf{Two-Stage Solution Framework: } We introduce a two-stage L2O inference framework.
    In the first stage, PDHG-Net delivers high-quality solutions. In the second stage, we further optimize these solutions by using a PDLP solver. This framework effectively accelerates the solving process while achieving relatively high precision.
    
    \item \textbf{Empirical Results: }  We conduct extensive experiments that showcase the superior performance of our framework in large-scale LP. 
    \textbf{Remarkably, our framework achieves up to a 3$\times$ speedup compared to PDLP.}
    Furthermore, we carry out additional experiments aimed at elucidating the mechanisms behind the acceleration, potentially providing insights for future L2O design.
\end{itemize}
\section{ PDHG-Net for large-scale LP}\label{sec:pdhg}

\subsection{Standard PDHG Algorithm}
Consider the LP problem $\mathcal{M} = (G; l,u,c; h)$ in standard form
\begin{equation}
\begin{aligned}
\min\ & c^\top x \\
\text{s.t.}\ 
& Gx \geq h \\
& l \leq x \leq u
\end{aligned}
\end{equation}
where $G \in \mathbb{R}^{m \times n}, h \in \mathbb{R}^m,  c \in \mathbb{R}^{n}, l \in(\mathbb{R} \cup\{-\infty\})^{n}, u \in(\mathbb{R} \cup\{+\infty\})^{n}$. Equivalently, we can solve the corresponding saddle point problem:
\vspace{-2mm}
\begin{equation}\label{lp_minmax}
\min_{l\leq x\leq u} \max_{y \geq 0} L(x,y; \mathcal{M}) = c^\top x - y^\top Gx + h^\top y
\end{equation}
\vspace{-5mm}

Primal-Dual Hybrid Gradient (PDHG) method, also referred to as the Chambolle-Pock algorithm \cite{chambolle2011first} in optimization literature, is a highly effective strategy for addressing the saddle point problem as outlined in (\ref{lp_minmax}). PDHG algorithm alternately updates the primal variable $x$ and the dual variable $y$ to find the saddle point of the Lagrangian $L$. To be more specific, it has the updating rule as follows.

\vspace{-8mm}

\begin{align*}
\intertext{\centering\textbf{Primal-Dual Hybrid Gradient (PDHG)}}
& - \text{Initialize}\ x^0 \in \mathbb{R}^n ,\ y^0 \in \mathbb{R}^m \\
& \text{For}\ \ k = 0, 1, 2,...,  K-1 \\
& \hspace{1em}\left\lfloor
\begin{aligned}
& x^{k+1} = \mathrm{\bf{Proj}}_{l\leq x\leq u} (x^k - \tau (c - G^\top y^{k})); \\
& y^{k+1} = \mathrm{\bf{Proj}}_{y\geq 0} (y^k + \sigma (h - 2 G x^{k+1} + G x^{k})).
\end{aligned}
\right. 
\end{align*}

\vspace{-4mm}


\begin{figure*}[t]
    \centering
    \includegraphics[width=0.85\textwidth]{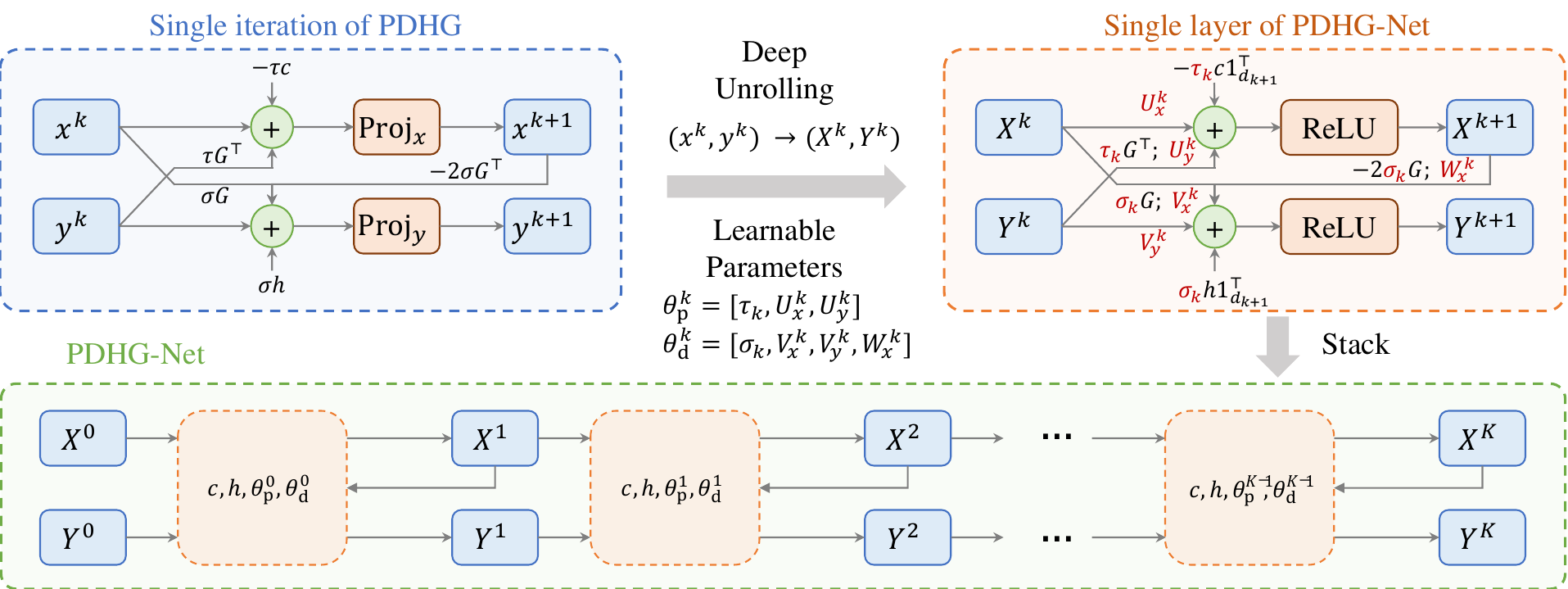}
    \caption{Overview of how each layer in PDHG-Net corresponds to each iteration of the traditional PDHG algorithm, along with the overall architecture of PDHG-Net}
    \vspace{-5mm}
    \label{overview}
\end{figure*}

Under regular assumptions, the PDHG algorithm has an ergodic convergence rate of $\mathcal{O}(1/K)$.

\begin{proposition}[\citet{chambolle2016ergodic}]\label{pock_proposition}
    Let ${(x^k, y^k)}_{k \geq 0}$ be the primal-dual variables generated by the PDHG algorithm for the LP problem $\mathcal{M} =  (G; l,u,c; h)$. If the step sizes $\tau, \sigma$ satisfy $\tau \sigma \|G\|_2^2 < 1$, then for any $(x,y) \in \mathbb{R}^n \times \mathbb{R}_{\geq 0}^m$ satisfying $l \leq x \leq u$, the primal-dual gap satisfies
    \begin{equation*}
    \begin{aligned}
        &\quad L(\bar{x}^k, y; \mathcal{M}) - L(x, \bar{y}^k; \mathcal{M}) \\ &\leq \frac{1}{2k} \Big( \frac{\|x-x^0\|^2}{\tau} + \frac{\|y-y^0\|^2}{\sigma} - (y - y^0)^\top G (x-x^0) \Big),
    \end{aligned}
    \end{equation*}
    where $\bar{x}^k = (\sum_{j=1}^k x^j) / k$, $\bar{y}^k = (\sum_{j=1}^k y^j) / k$, and $L$ is the Lagrangian defined by (\ref{lp_minmax}).
\end{proposition}


\subsection{Design of PDHG-Net}

In this part, we describe how we design
PDHG-Net by unrolling PDHG algorithm
and incorporating techniques from graph neural networks. 

In the PDHG algorithm, each iteration involves period-2 updates of primal and dual variables, as illustrated in Figure \ref{overview}. In particular, at the $k$-th iteration, the algorithm first gets the primal variable \( x^{k+1} \) starting from \( (x^k, y^k) \), and then gets the dual variable \( y^{k+1} \) using the newly obtained \( x^{k+1} \) along with \( y^k \). 
We replace the primal-update and dual-update step with neural network (NN) blocks parameterized by \( \Theta^{k}_{\rm p} \) and \( \Theta^{k}_{\rm d} \), respectively.






To provide insights for the design of PDHG-Net blocks, we focus on two components of the PDHG iteration: the proximal operator and the linear combination of the primal-dual variables. These two components respectively lead us to incorporate two distinct approaches in our design: ReLU activation and channel expansion.

\vspace{-2mm}

\paragraph{Proximal operator -- ReLU activation.} The proximal operators, specifically the projection functions $\mathrm{\bf{Proj}}_{y\geq 0}$ and $\mathrm{\bf{Proj}}_{l\leq x\leq u}$, exhibit a piece-wise linear behavior on each entry. This observation naturally leads to the use of ReLU activations in our network. 
In the forthcoming Theorem \ref{alignment} and its proof, we will see that this property is useful to help prove that a PDHG-Net can accurately replicate the PDHG algorithm.

\vspace{-2mm}

\paragraph{Channel expansion.} 
In this part, we focus on the primal updates as an illustrative example and the analysis of the dual updates can be derived similarly. We expand the $n$-dimensional vectors $x^k, y^k$ into $(n \times d_{k})$-dimensional matrices $X^k, Y^k$ with $d_k$ columns (or called channels following the convention of neural network literature).
The linear combination $x^k - \tau(c - G^{\top} y^k)$ of primal-dual is replaced by $X^k U^k_x - \tau_k(c \cdot \mathbf{1}^\top_{d_{k+1}} - G^\top Y^k U^k_y)$ where $\Theta^k_{\rm p} = (\tau_k, U^k_x, U^k_y) \in \mathbb{R} \times \mathbb{R}^{d_k \times d_{k+1}} \times \mathbb{R}^{d_k \times d_{k+1}}$ is the trainable parameter of the $k$-th primal NN block. This design is for the network to generalize to LP instances of different sizes and enhance the expressivity.
\begin{itemize}[leftmargin=*]
    \vspace{-2mm}
    \item \textbf{Generalizability to LP instances of different sizes}: It is important to note that our NN design diverges from classical deep unrolling for sparse coding problems, such as the unrolled/unfolded iterative shrinkage thresholding algorithm (ISTA) \cite{gregor2010learning}.
    Following the principle of classical unrolling,
    a natural idea would be to unroll
     \( x^k - \tau(c - G^{\top} y^k) \)
    to 
    \begin{equation}\label{eq::scale}
    x^k - \tau(c -  W^k  y^k) 
    \end{equation}
    where $W^k$ is trainable matrix. 
    However, 
    for different LP problems, the sizes 
    of the constraint matrices \( G \) vary, 
    so the replaced trainable matrix can only have a fixed size of \(n \times m\), rendering it unsuitable for applying to LP problems with different sizes. 
    Note that this issue does not exist in ISTA, 
    since the transformation matrix in the ISTA algorithm is solely related to the dictionary matrix, which remains fixed in sparse coding problems. 
    To address the generalizability issue of using the component \eqref{eq::scale} in neural nets,
    we expand \( x^k, y^k \) into \( X^k,Y^k \) with multiple channels and apply a trainable matrix on their right. 
    This method, which is motivated by GNN, makes PDHG-Net
    generalizable to LP problems with different sizes.

    \vspace{-2mm}
    \item \textbf{Expressivity}: 
   Intuitively, increasing the network width, i.e. the number of channels $d_k$ in our setting, can enhance the network's expressivity. 
  As we demonstrate in the subsequent Theorem \ref{alignment}, if $d_k$'s are larger than a certain fixed constant (e.g. 10), PDHG-Net possesses sufficient expressive power to accurately recover iterations of PDHG algorithms. 
    
\end{itemize}

\vspace{-2mm}


After we design each NN block, we combine the $K$ primal and $K$ dual blocks by the same pattern that the classical PDHG algorithm forms. Notably, as depicted in Figure \ref{overview}, the architecture resembles the form of a GNN, updating primal and dual variables in a period-2 manner. There are at least two major differences between PDHG-Net and the common GNN architectures designed for solving LP-based problems \cite{conf/nips/GasseCFCL19, chen2022representing}: first, PDHG-Net employs a period-2 update cycle, contrasting with the period-1 update cycle in GNN. Second, within PDHG-Net's update period, dual iterations undergo twice as many aggregations compared to primal iterations, while variable and constraint nodes share the same number of aggregations in GNN's update period. 

Based on the analysis above, we propose the architecture of PDHG-Net formally below. 

\vspace{-10mm}

\begin{align*}
\label{Forward Propagation of PDHG-Net}
& \intertext{\centering\textbf{Architecture of PDHG-Net}} 
& - \text{Initialize}\ X^0 = [x^0, l, u, c],\ Y^0 = [y^0, h]\\
& \text{For}\ \ k = 0, 1, 2,...,  K-1 \\
& \hspace{1em}\left\lfloor
\begin{aligned}
X^{k+1} &= \operatorname{ReLU} \big( X^k U^k_x - \tau_k(c \cdot \mathbf{1}^\top_{d_{k+1}} - G^\top Y^k U^k_y) \big), \\
Y^{k+1} &= \operatorname{ReLU} \big(Y^k V^k_y \\
&\ \  + \sigma_k(h \cdot \mathbf{1}^\top_{d_{k+1}}- 2G X^{k+1} W^{k}_x + G X^k V^k_x ) \big),
\end{aligned}
\right. \\
& - \text{Output}\ X^K \in \mathbb{R}^{n} ,\ Y^K \in \mathbb{R}^{m}
\end{align*}
The trainable parameter is $\Theta = \{\Theta^k_{\rm p}, \Theta^k_{\rm d}\}_{k=0}^{K-1}$, where $\Theta^k_{\rm p} = (\tau_k, U^k_x, U^k_y) \in \mathbb{R} \times \mathbb{R}^{d_k \times d_{k+1}} \times \mathbb{R}^{d_k \times d_{k+1}}$ and $\Theta^k_{\rm d} = (\sigma_k, V^k_x, V^k_y, W^k_x) \in \mathbb{R} \times \mathbb{R}^{d_k \times d_{k+1}} \times \mathbb{R}^{d_k \times d_{k+1}} \times \mathbb{R}^{d_{k+1} \times d_{k+1}}$. 



The following theorem reveals the capability of PDHG-Net to reproduce the classical PDHG algorithm. It indicates an adequate expressive power of PDHG-Net.



\begin{theorem}\label{alignment}
Given any pre-determined network depth $K$ and the widths $\{d_k\}_{k\leq K-1}$ with $d_k \geq 10$, there exists a $K$-layer PDHG-Net with its parameter assignment $\Theta_{\rm PDHG}$ satisfying the following property: given any LP problem $\mathcal{M} = (G; l, u, c; h)$ and its corresponding primal-dual sequence $(x^k, y^k)_{k \leq K}$ generated by PDHG algorithm within $K$ iterations, we have
\begin{enumerate}
    \item For any hidden layer $k$, both $\bar{x}^k$ and $x^k$ can be represented by a linear combination of $X^k$'s channels, both $\bar{y}^k$ and $y^k$ can be represented by a linear combination of $Y^k$'s channels. Importantly, these linear combinations do not rely on the LP problem $\mathcal{M}$.
    
    
    \item PDHG-Net's output embeddings $X^K \in \mathbb{R}^{n \times 1}$ and $Y^K \in \mathbb{R}^{m \times 1}$ are equal to the outputs $\bar{x}^K$ and $\bar{y}^K$ of the PDHG algorithm, respectively.
\end{enumerate}
\end{theorem}

Theorem \ref{alignment} provides a guarantee of an \textit{exact} alignment between a finite-width PDHG-Net and the standard PDHG algorithm.
Combining Proposition \ref{pock_proposition} and Theorem \ref{alignment}, we propose an upper bound of the required number of neurons to achieve an approximation error of $\epsilon$.


\begin{theorem}\label{poly expressive thm}
Given the approximation error bound $\epsilon$, there exists a PDHG-Net with $\mathcal{O}(1/\epsilon)$ number of neurons and the parameter assignment $\Theta_{\rm PDHG}$ fulfilling the following property. For any LP problem $\mathcal{M} = (G;l,u,c;h)$ and $(x,y) \in \mathbb{R}^n \times \mathbb{R}_{\geq 0}^m$ satisfying $l \leq x \leq u$, it holds that
\begin{equation*}
    L(X^K, y; \mathcal{M}) - L(x, Y^K; \mathcal{M}) < \epsilon.
\end{equation*}
\end{theorem}
\vspace{-2mm}

As previously mentioned, our PDHG-Net is closely related to GNN, but also has some differences due to the unrolling nature.
We explain the benefit of our design
from a theoretical perspective below. 

A recent work \citet{chen2022representing} proves that GNNs possess strong expressive power to represent the optimal solutions of LP problems. 
The motivation is to understand the power of
GNNs commonly used in L2O for optimization problems. 
The representation power result is rather non-trivial,
since the expressive power of GNNs is well-known to be limited by the Weisfeiler-Lehman (WL) test \cite{xu2018powerful}. \citet{chen2022representing} ingeniously proves that this limitation does not pose an issue for LP, as those LP problems indistinguishable by WL-test are shown to share the same optimal solutions if exist. 

Compared to \citet{chen2022representing},
our proof of expressive power of PDHG-Net has two advantages.
First, the proof is rather concise. 
This is because PDHG-Net is designed by unrolling the classical PDHG algorithm whose convergence has been confirmed,
in contrast to the original design of GNNs (e.g. \cite{gasse2019exact}) which
just applies an existing architecture to optimization 
problems without much theoretical justification. 
Thus, the proof of Theorem \ref{poly expressive thm} does not rely on analyzing the separation power of WL-test in LP problems, making it more concise.
Second, perhaps more importantly,
under a given error tolerance $\epsilon$, we demonstrate that the number of neurons required is upper-bounded by $\mathcal{O}(1/\epsilon)$ in PDHG-Net. In contrast, \citet{chen2022representing}, while utilizing the universal approximation properties of MLPs, does not explore (at least not explicitly) the relationship between network size and expressive power.

\subsection{Network Training}

\vspace{3pt}\noindent \textbf{Training dataset.}
The training dataset is a set of instances denoted by $\mathcal{I} = \{( \mathcal{M}, z^*) \}$. The input of an instance is an LP problem $\mathcal{M} = (G; l, u, c; h)$; the label $z^* = (x^*, y^*)$ is its corresponding near-optimal primal-dual solution, which is acquired by adopting a well-established solver such as PDLP with a given stopping criterion.


\vspace{3pt}\noindent \textbf{Loss Function.}\label{LossFunctionDesign}
Given the parameter $\Theta = \{\Theta^k_{\rm p}, \Theta^k_{\rm d}\}_{k=0}^{K-1}$ and the input LP problem $\mathcal{M}$, we denote the output $(X^K, Y^K)$ of the PDHG-Net by $Z^K(\mathcal{M}; \Theta)$. We seek to reduce the distance between $Z^K(\mathcal{M}; \Theta) = (X^K, Y^K)$ and $z^* = (x^*, y^*)$ for each training LP problem. Therefore, we train the PDHG-Net to minimize the $\ell_2$ square loss
\begin{equation}\label{eq: LossFunction}
\begin{aligned}
    \min_{\Theta} \mathcal{L}_{\mathcal{I}}(\Theta)=\frac{1}{|\mathcal{I}|} \sum_{(\mathcal{M}, z^*) \in \mathcal{I}} \left\| Z^K(\mathcal{M}; \Theta) - z^* \right\|_2^2.
\end{aligned}
\end{equation}

Theorem \ref{alignment} indicates the existence of model $\Theta_{\rm PDHG}$ that perfectly replicates the PDHG algorithm. This naturally leads us to the conclusion that a well-trained PDHG-Net may speed up the convergence to optimal solutions.


\begin{corollary}
If $\Theta_{\rm PDHG}$ is not the global minima of the loss function $\mathcal{L}_{\mathcal{I}}$, a $K$-layer PDHG-Net can potentially achieve primal-dual solutions closer to the optimal ones than the solutions generated by the standard PDHG algorithm implemented for $K$ iterations.
\end{corollary}

Through the training process, our goal is to find the parameter $\Theta_0$ satisfying
\begin{equation*}
    \mathcal{L}_{\mathcal{I}}(\Theta_0) \approx \min_{\Theta} \mathcal{L}_{\mathcal{I}}(\Theta) < \mathcal{L}_{\mathcal{I}}(\Theta_{\rm PDHG}).
\end{equation*}

\subsection{Two-Stage Framework}
\begin{figure}[t]
    \centering
    \includegraphics[scale=0.45]{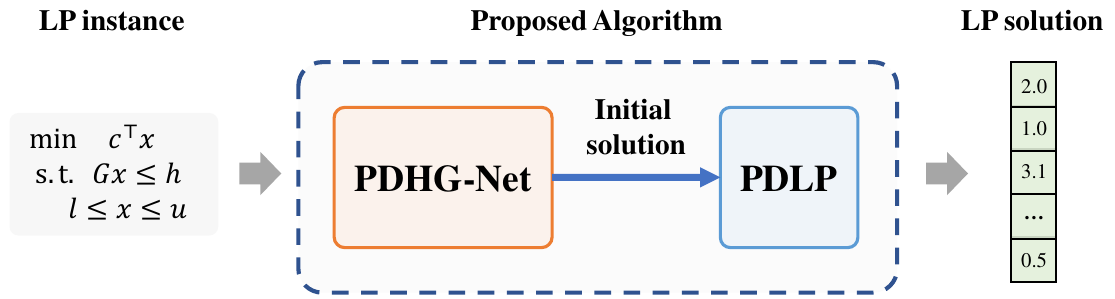}
    \vspace{-2mm}
    \caption{The proposed post-processing procedure warm-starts the PDLP solver using the prediction of PDHG-Net as initial solutions to ensure optimality.}
    \vspace{-3mm}
    \label{ee}
\end{figure}
\vspace{-2mm}

To enhance the solution precision, we propose a two-stage framework that first obtains the predicted solutions, and then incorporates a post-processing step to refine these predictions and ensure optimality. 
For this purpose, we select the PDLP solver from Google's Ortools~\cite{ortools}, renowned for its exemplary implementation of the PDHG algorithm, as the key element in our post-processing module. 
As demonstrated in Figure~\ref{ee}, this module involves using the predicted solutions to warm-start the PDLP solver, thereby improving the efficiency in solving the original problem.
Specifically, given an LP instance, we first feed it into the well-trained PDHG-Net to acquire the prediction of primal and dual solutions $\{X^K,Y^K\}$.
Then, we warm-start PDLP by setting $x^0 = X^K$ and $y^0 = Y^K$, and let the solver iteratively update primal and dual solutions, until the termination criteria are met.
For comprehensive details on the implementation, including features and training settings, please see Appendix~\ref{app:feat}.

\section{Implementation Settings}
\textbf{Datasets.} 
We conduct comprehensive experiments on a selection of representative LP instances grounded in real-world scenarios. 
Specifically, our experiments involve the PageRank dataset for large-scale LP problems, as well as the Item Placement (\texttt{IP}) and the Workload Appropriation (\texttt{WA}) datasets~\cite{gasse2022machine} for complex LP relaxations.
Particularly, the PageRank dataset is acquired by generating instances in accordance with the methodology outlined in~\cite{applegate2021practical}.
For the \texttt{IP} and \texttt{WA} datasets, the generation protocol is adjusted to ensure that each instance presents a significant computational challenge.
To ensure broad applicability, each benchmark dataset encompasses instances across a spectrum of problem sizes. 
Detailed information on the generation settings and specific problem sizes is available in the Appendix \ref{app:dataset}. 

\textbf{Software and Hardware settings.} 
For the performance comparisons, we employ the off-the-shelf implementation of PDLP from Google's OrTools 9.8.3296~\cite{ortools}.
Gurobi 10.0.2's LP solvers~\cite{gurobi} are also employed for experiments on the PageRank dataset.
Our proposed framework was implemented utilizing PyTorch version 2.1.1.~\cite{paszke2019pytorch}, and all computational studies are carried out on a cloud server with one NVIDIA Tesla V100-32GB GPU, one Intel Xeon Platinum 8280 CPU and 3,072 GB RAM.
For clarity and simplicity in subsequent discussions, we will refer to the performance of PDLP as 'PDLP' and that of our proposed framework as 'ours'.
To assess the performance of our proposed framework, we calculate the improvement ratio over PDLP using the following equation:
\begin{equation*}
    \text{Improv.} = \frac{\text{PDLP}-\text{ours}}{\text{PDLP}},
\end{equation*}
where this metric is applicable to both the solving time and the number of iterations. 

\textbf{Training.} 
Considering the constraints of computing resources and the need for computational efficiency, we utilize a 4-layer PDHG-Net to strike an optimal balance between network performance and inference costs. 
Given the training dataset $\mathcal{I}=\{(\mathcal{M}_1,z^*_1),\dots,(\mathcal{M}_{|\mathcal{I}|},z^*_{|\mathcal{I}|})\}$, we extract key information and encode it into each $\mathcal{M}_i$ to form $\Bar{\mathcal{M}}_i$, as detailed in Appendix~\ref{app:feat}. The dataset $\{(\Bar{\mathcal{M}}_1,z^*_1),\dots,(\Bar{\mathcal{M}}_{|\mathcal{I}|},z^*_{|\mathcal{I}|})\}$ is then split into training and validation sets with a $9:1$ ratio.
The training process entailed minimizing the $\ell_2$ square loss, as described in Equation~\ref{eq: LossFunction}, and updating the trainable parameter $\Theta$ through backward propagation, where we utilize an Adam optimizer with a learning rate of $10^{-4}$. 
After a certain number of iterations, the model exhibiting the lowest validation loss was selected as the output model.  



\section{Numerical Experiments}

In this section, we embark on a detailed empirical examination of our proposed PDHG-based two-stage framework's effectiveness in managing complex linear programs. 
Initially, we showcase the framework's performance in tackling large-scale LP problems. 
We then showcase the framework's capability to produce solutions for LP instances, which typically entail considerable convergence time for FOMs.
Following this, we conduct a series of experiments aimed at uncovering the underlying factors contributing to the PDHG-Net's remarkable performance. 
Finally, a detailed investigation of experiments is undertaken to assess the framework's generalization power and scalability.
For reproducibility, our code can be found at \url{https://github.com/NetSysOpt/PDHG-Net.git}.

\setlength{\tabcolsep}{4.1mm}{
\begin{table}[h]
\centering
\caption{Solve time comparison between the proposed framework and vanilla PDLP on PageRank instances. The improvement ratio of the solving time is also reported.}
\label{t:pagerank}
\renewcommand\arraystretch{0.7}
\begin{tabular}{cccc}
\toprule
     \# nodes &ours&PDLP & Improv.    \\
\midrule
$10^3$  &0.01sec. &0.04sec.&     $\uparrow 45.7\%$ \\
$10^4$  &0.4 sec. &1.1sec.&                             $\uparrow 67.6\%$ \\
$10^5$  &22.4sec. &71.3sec.&                            $\uparrow 68.6\%$ \\
$10^6$  &4,508sec.&16,502sec.&                         $\uparrow 72.7\%$ \\
\bottomrule
\end{tabular}
\end{table}
\vspace{-3mm}
}

\subsection{Comparing against vanilla PDLP on large-scale LP problems}
In this section, we illustrate how our proposed framework improves the efficiency of PDLP in managing large-scale LP problems through comprehensive experiments conducted on the PageRank dataset across a range of sizes.
Specifically, we benchmark our approach against the vanilla PDLP solver, and report the solving time along with the improvement ratio brought by our framework. 

Detailed numerical results are presented in Table~\ref{t:pagerank}. 
A comparative analysis of the solving time reveals a notable efficiency of our proposed framework: on average, it solves the PageRank problem $64.3\%$ faster than PDLP. 
Importantly, as the problem size increases, our proposed framework not only sustains a consistent level of performance but also demonstrates a steadily increasing improvement in efficiency.
Particularly, our approach consistently outperforms standard PDLP on problems with more than $10^5$ nodes, solving instances up to three times faster.
Such observations not only underscore the inherent efficacy of PDLP in managing large-scale LP problems, but also highlight the significant enhancements our framework brings to these capabilities.

The significant performance improvement between our proposed framework and the vanilla PDLP can be elucidated by two key factors:
\textbf{(1)} PDLP typically begins the solving process by initializing both primal and dual solutions as zero vectors. 
This approach overlooks the distinct characteristics of the specific LP instance. 
Conversely, our model strategically utilizes the unique aspects of the problem, generating initial points that are specifically tailored for the given instance.
\textbf{(2)} The initial solution proposed by our model closely approximates the optimal solution, which likely reduces the number of iterations needed to achieve optimality, thereby significantly hastening the solving process.
The collected data from these experiments distinctly demonstrate that our proposed method consistently outperforms both PDLP and Gurobi.

\subsection{Comparing against PDLP on difficult linear relaxations}
FOM-based methods, including PDLP, often experience slow convergence when applied to complex LP problems with stringent constraints.
The objective of this section is to demonstrate how our proposed approach can overcome these challenges by providing high-quality initial solutions. 
Our evaluation encompasses two datasets, featuring LP relaxations of challenging Mixed Integer Programs: \texttt{IP} and \texttt{WA}. 
We conduct a comparative analysis of our approach's performance against the default PDLP. 
It's important to note that the solving precision was set at $10^{-4}$ to accommodate PDLP's precision limitations, ensuring comparisons stay within its operational boundaries.

In Table~\ref{t:ipwa}, we comprehensively present our numerical experiments, comparing our proposed framework with traditional PDLP. 
The table is organized into two sections: the upper part focuses on solving time as the primary metric, while the lower part assesses the number of iterations required for solving the problem. 
The data from Table~\ref{t:ipwa} reveals significant efficiency enhancements of our framework in comparison to PDLP. 
In the \texttt{IP} dataset, our framework shows a notable improvement of $19.5\%$ in both solving time and the number of iterations for smaller instances, compared to PDLP. 
This enhancement becomes even more pronounced in larger \texttt{IP} problems, with a $21.7\%$  reduction in solving time and a $27.8\%$ decrease in the number of iterations. 
These results align with PDLP's growing proficiency in handling larger LP instances. 
Additionally, the performance of our framework on the \texttt{WA} dataset also consistently surpasses that of PDLP. 
Furthermore, an increase in efficacy is also observed as the size of the problems in the \texttt{WA} dataset grows, further demonstrating the capability of our approach.
Such consistency is evident across different instance sizes within both datasets, as detailed in Table~\ref{t:ipwa}, indicating that our model's efficiency is robust and not negatively impacted by variations in instance size.
These findings underscore our proposed framework's ability to effectively tackle complex problems that are typically formidable for conventional FOMs. 



\vspace{-2mm}

\setlength{\tabcolsep}{0.8mm}{
\begin{table}[htbp]
\renewcommand\arraystretch{0.7}
\centering
\caption{Results of comparing the proposed framework against default PDLP in solving \texttt{IP} and \texttt{WA} instances. The top and bottom sections present the solving time and number of iterations, respectively, where bold fonts represent better performance.}
\label{t:ipwa}
\begin{tabular}{c ccc ccc}
\toprule
        \multirow{2}{*}{dataset.}& \multicolumn{3}{c}{time (sec.)}& \multicolumn{3}{c}{\# iters.}\\
        \cmidrule(l){2-4}\cmidrule(l){5-7}
             &ours & PDLP &  Improv.&ours & PDLP & Improv.\\
\midrule
 \texttt{IP-S}& \small \textbf{9.2}& \small 11.4 & \small $\uparrow 19.5\%$ & \small \textbf{422}&\small 525 & \small $\uparrow 19.5\%$\\
 \texttt{IP-L}&\small \textbf{7,866.3}&\small  10,045.6&\small $\uparrow 21.7\%$& \small \textbf{6,048}&\small  8,380&\small $\uparrow 27.8\%$\\
\midrule\midrule
 \texttt{WA-S}& \small \textbf{114.7}&\small  137.8 &\small  $\uparrow 16.7\%$& \small  \textbf{8,262} & \small 9,946 &  \small $\uparrow 16.9\%$ \\
 \texttt{WA-L}& \small \textbf{4817.6}&\small  6426.2&\small  $\uparrow 25.0\%$& \small \textbf{14,259}&\small 17,280& \small  $\uparrow 17.5\%$ \\

\bottomrule
\end{tabular}
\end{table}
}

\vspace{-4mm}

\subsection{Understandings of why PDHG-Net works}
In this section, we explore the underlying factors of how our proposed framework enhances the solving of LP problems through experiments. Specifically, we address the following questions:

\textbf{Q1: Can the proposed PDHG-Net imitate the PDHG algorithm?}

\textbf{Q2: Why is our method capable of accelerating the solving process of PDLP solver?}


\begin{figure}
    \centering
    \includegraphics[scale=0.32]{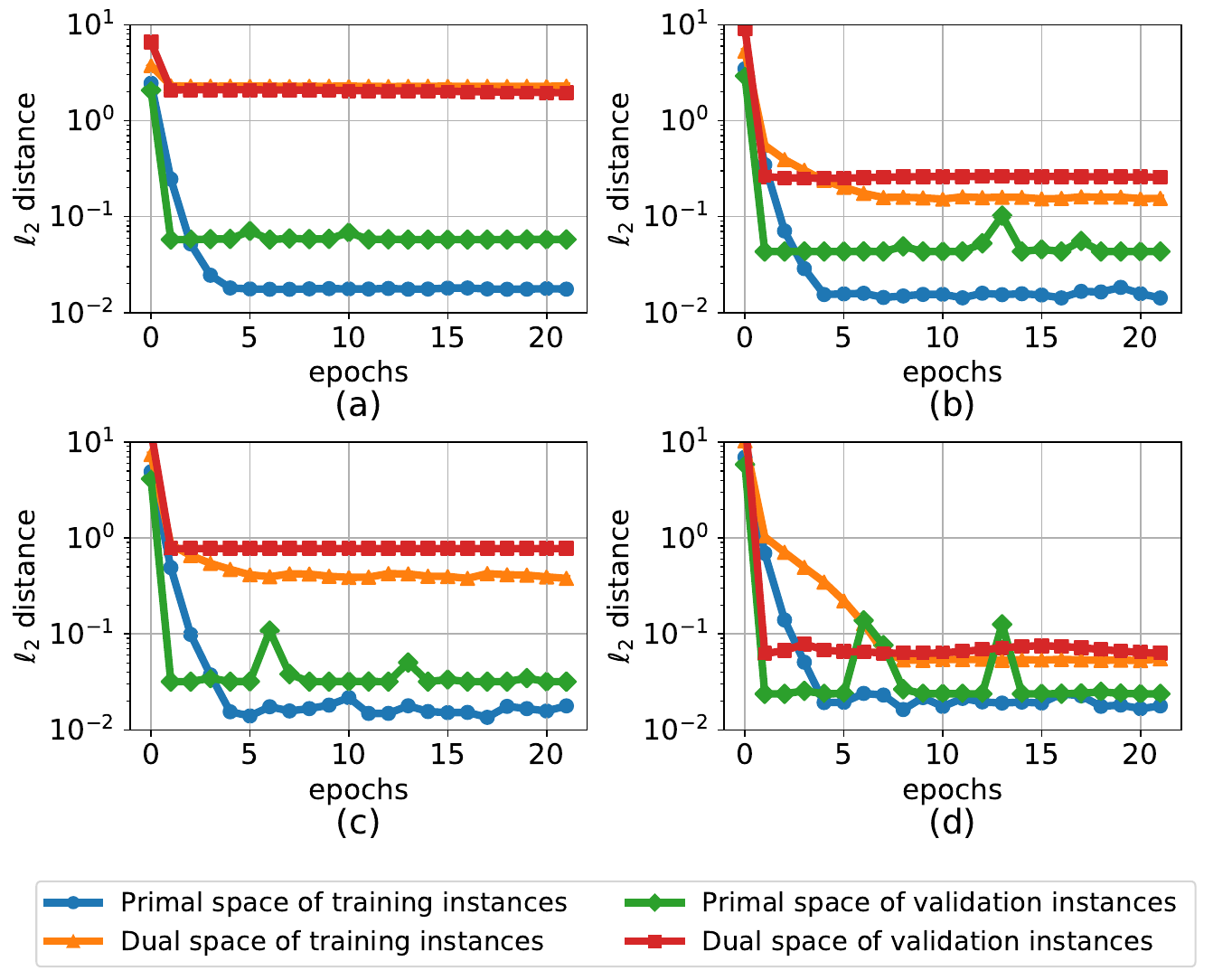}
    \vspace{-1mm}
    \caption{The distance between the predicted solution of PDHG-Net and optimal solution in PageRank training and validation instances with (a) $5\times10^3$, (b) $1\times10^4$, (c) $2\times10^4$, (d) $4\times10^4$ variable sizes.}
    \label{train test loss}
    \vspace{-1mm}
\end{figure}

\textbf{Align with the PDHG (Q1).}
In Figure \ref{train test loss}, we report the average $\ell_2$ distance \footnote[1]{During the network's training phase, the $\ell_2$ square loss is employed as the loss function. For the evaluation of experimental results, we measure the $\ell_2$ distance between the predicted and optimal solutions, which corresponds to the square root of the $\ell_2$ square loss.} between the predicted solution $Z^K({\rm \mathcal{M}}; \Theta) = (X^K, Y^K)$ and the obtained optimal solution $z^* = (x^*, y^*)$ for training and validation sets during the training process. 

Our experimental results reveal that the solutions inferred by PDHG-Net increasingly converge to the optimal solutions throughout training, ultimately achieving a markedly small discrepancy for both primal and dual solutions. 
This consistency underscores that the proposed PDHG-Net aligns well with the PDHG algorithm.
Furthermore, the validation $\ell_2$ distances, maintaining the same order of magnitude across unseen instances, attest to our model's robust generalization ability.



\begin{figure}[t]
\centering
\subfigure[Solving time]{
\begin{minipage}[t]{0.42\linewidth}
\includegraphics[scale=0.29]{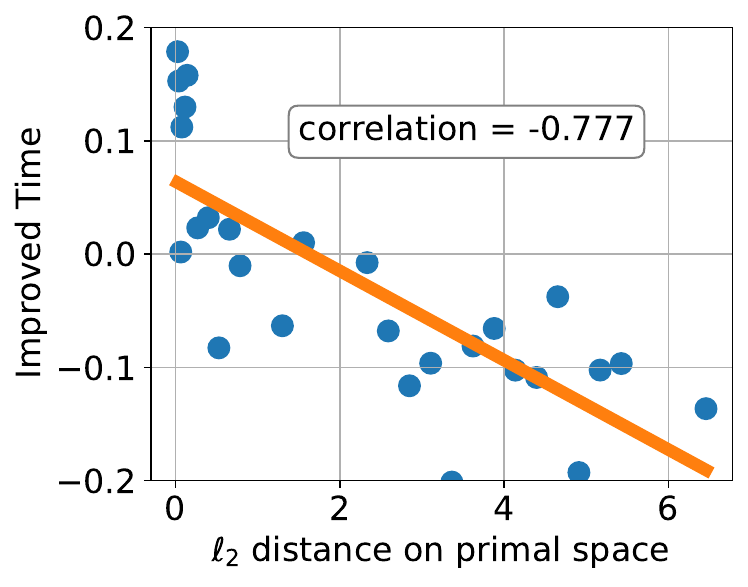}
\end{minipage}%
}%
\hfill
\subfigure[Number of iterations]{
\begin{minipage}[t]{0.45\linewidth}
\centering
\includegraphics[scale=0.29]{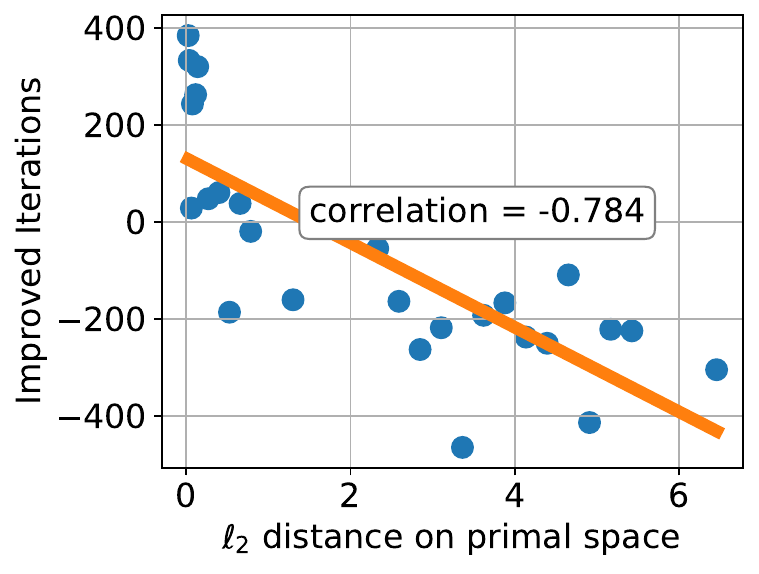}
\end{minipage}%
}%
\centering
\caption{We present the improvement ratio in both solving time and the number of iterations for solutions extrapolated at varying distances from the optimal solution. Each blue dot symbolizes an extrapolated solution, while the yellow line represents the trend line fitted through these points. Results demonstrate a strong correlation.}
\label{distance_time}
\end{figure}


\textbf{Accelarating PDLP solving process (Q2).} 
To explore the mechanisms behind our framework's improved solving efficiency, we examine the PDHG-Net on the PageRank dataset. 
By extrapolating PDHG-Net's output $Z^K({\rm \mathcal{M}}; \Theta) = (X^K, Y^K)$, from the collected optimal solution $z^* = (x^*, y^*)$, we generate solutions that vary in their proximity to $z^*$. 
In Figure \ref{distance_time}, we illustrate the improvement ratio of warm-starting PDLP with these solutions in solving time and numbers of iterations compared against the vanilla PDLP solver. 
The findings demonstrate a strong correlation between solving performance and the extent of extrapolation, which is consistent with Proposition~\ref{pock_proposition}. 
Therefore, the proposed PDHG-Net accelerates the solving process by providing high-quality initial solutions.

In addition, we evaluate our framework's efficiency through the number of restarts required to solve the problem. 
The vanilla PDLP solver performs restarts under specific conditions to enhance performance and prevent stalling~\cite{lu2023practical}.
By comparing our framework to the standard PDLP solver on PageRank instances of differing sizes, we present the number of restarts during the solution process for both approaches in Table \ref{restart}.
The findings indicate a significant decrease in the number of restarts when PDLP is warm-started with predictions from PDHG-Net.
This suggests that our framework enables a more efficient optimization trajectory, thus accelerating the problem-solving process.



\setlength{\tabcolsep}{1.2mm}{
\begin{table}[h]
\centering
\renewcommand\arraystretch{0.6}
\caption{The average number of restarts in the PDLP solving process with our framework (ours) and default settings (represented by PDLP). }
\label{restart}
\begin{tabular}{cc cccc}
\toprule
\multicolumn{2}{c}{\# of Nodes} & $5\times 10^3$ & $1\times 10^4$ & $2\times 10^4$ & $4\times 10^4$ \\
\midrule\midrule
\multirow{2}{*}{\# restarts}&Ours                                      & 2.2            & 4.15           & 2.0            & 2.0            \\
\cmidrule(l){2-6}
&PDLP                                      & 5.9            & 11.7           & 20.25          & 11.3           \\ \bottomrule
\end{tabular}
\end{table}
}

Therefore, we conclude that our framework markedly enhances the solving process by warm-starting the solver with high-quality solutions and facilitating a more efficient gap-closing approach.

\subsection{Generalization on different sizes}

\setlength{\tabcolsep}{1.5mm}{
    \begin{table}[t]
    \renewcommand\arraystretch{0.7}
    \centering
    \caption{Solving time and number of iterations for PageRank, \texttt{IP} and \texttt{WA} instances larger than training set sizes. For clarity, we denote the size of the largest instance of \texttt{IP} and \texttt{WA} datasets in Table~\ref{t:ipwa} as \texttt{Large}.}
    \label{t:gen}
    \begin{tabular}{c c c| rrr}
    \toprule
           metric &Dataset &size  &ours &PDLP &Improv. \\
    \midrule
   \small\multirow{4}{*}{time (sec.)}&\multirow{2}{*}{\small PageRank} & \small $5\times10^{4}$ & \small\textbf{ 5.5} & \small 11.2 & \small$\uparrow 50.9\%$\\
     && \small $1\times10^{5}$ & \small\textbf{17.0}& \small 32.5& \small$\uparrow47.8\%$\\
    \cmidrule(l){2-6}
  &\small \texttt{IP}  &  \small \texttt{Large} & \small\textbf{6796.7}& \small 8631.4& \small$\uparrow  21.3\%$\\
    \cmidrule(l){2-6}
  &\small \texttt{WA}  &  \small \texttt{Large} & \small\textbf{ 5599.1}& \small 5859.4& \small$\uparrow  4.4\%$\\

    \midrule\midrule
  \small \multirow{4}{*}{\# iter.}&\multirow{2}{*}{\small PageRank} & \small $5\times10^{4}$ & \small\textbf{1,605} & \small 3,397&\small$ \uparrow 52.7\%$\\
    & & \small $1\times10^{5}$ & \small\textbf{1,958} & \small3,914&\small$\uparrow 50.0\%$  \\
    \cmidrule(l){2-6}
  &\small \texttt{IP}  &  \small \texttt{Large} & \small\textbf{7,291} & \small 8,970&\small$ \uparrow 18.7\%$\\
    \cmidrule(l){2-6}
  &\small \texttt{WA}  &  \small \texttt{Large}&\small\textbf{16,166} & \small 17,280&\small$ \uparrow 6.4\%$\\
    \bottomrule
    \vspace{-5mm}
    \end{tabular}
    \end{table}
}

We investigate the capability of the proposed method to generalize beyond the scope of its training set, focusing on its adaptability to larger problems. 
For this purpose, we utilize models trained on the PageRank, \texttt{IP}, and \texttt{WA} datasets to generate initial solutions for problems of larger scale.

We present a comprehensive overview of our model's generalization capabilities in Table~\ref{t:gen}, where we compare its solving time and number of iterations against those of the standard PDLP.  
This comparative analysis underscores that our model consistently surpasses vanilla PDLP, registering an average acceleration of $31.1\%$ across these datasets.
Notably, in the PageRank dataset, our model achieves performance levels comparable to those models trained on datasets of identical size. 
Similarly, this performance improvement is evident when considering the metric of the number of iterations. 
As indicated in the bottom section of Table~\ref{t:gen}, our model, on average, reduces the number of required iterations to reach optimality by approximately $31.8\%$.
These observations collectively underscore the robustness of our proposed framework, particularly in its ability to generalize effectively across a diverse range of problem sizes. 
On the other hand, the observed consistency in performance may be attributed to the model's proficiency in identifying and utilizing common features inherent in a series of similar LP instances. 

\subsection{Scalability of PDHG-Net}
In this section, we evaluate the scalability of our proposed framework by analyzing the proportion of PDHG-Net inference time within the entire solving process. 
We gathered both the inference (GPU time) and the solving time (CPU time) for PDLP after warm-starting it with predictions given by PDHG-Net on PageRank instances of varying sizes.
To simplify our analysis, we introduce a new metric termed 'ratio' to quantify the proportion of GPU time in relation to the total evaluation time, which is defined as $\text{ratio}=\frac{\text{GPU time}}{\text{Search time}}$.
The findings in Table~\ref{t:gputime} reveal that PDLP-Net's inference time constitutes only a minor fraction of the total solving time. 
Notably, as the size of the dataset's variables increases, the share of GPU time for PDLP-Net inference markedly diminishes. 
This trend underscores our method's efficiency in scaling up to large-scale datasets, rendering the GPU time increasingly insignificant.


\setlength{\tabcolsep}{2.1mm}{
    \begin{table}[t]
    \centering
    \renewcommand\arraystretch{0.6}
    \caption{Comparison of total solving time and GPU time for initial solutions, including the ratio of GPU time to total solving time.}
    \label{t:gputime}
    \begin{tabular}{ c| rrrr}
    \toprule
               \# nodes. & $10^3$& $10^4$ & $10^5$ & $10^6$ \\
    \midrule
     GPU time (sec.)& 0.01        & 0.02  & 0.21  &1.12\\
      \midrule
      CPU time (sec.)& 0.02    & 0.4  & 22.4  &4,508.3\\
      Ratio & $52.6\%$          & $5.7\%$  & $0.9\%$    &$0.02\%$\\
    \bottomrule
    \end{tabular}
    \label{portion_of_GPU}
    \end{table}
}

\subsection{Comparing PDHG-Net Against GNN}
In previous sections, we distinguish the PDHG-Net from Graph Neural Networks (GNN) based on network architectures.
Another compelling aspect involves a direct performance comparison of the proposed framework when implemented using these two networks. 
To ensure fairness, models generating the initial solutions were trained under identical settings.
In Table~\ref{t:gnn}, we present the numerical results on the PageRank dataset. 
The performance is evaluated in two dimensions: the improvement ratio in solving time relative to vanilla PDLP and the $\ell_2$ distance from the optimal solution. 
The results clearly indicate that the PDHG-Net implementation not only enhances solving efficiency but also achieves lower $\ell_2$ distance in predictions.
Moreover, the $\ell_2$ distance between GNN's predictions and optimal solutions deteriorates with increasing problem size.
This observation indicates that GNN lacks the inherent robustness of PDHG in managing instances of varied sizes, where the performance of PDHG-Net highlights a relatively stable performance on diverse problem dimensions.
This study demonstrates PDHG-Net's superior efficacy in addressing challenging linear programming compared to GNN.

\setlength{\tabcolsep}{3.2mm}{
    \begin{table}[t]
    \renewcommand\arraystretch{0.6}
    \centering
    \caption{Comparison of improvement ratio and $\ell_2$ distance between the proposed framework implemented with PDHG-Net and GNN. }
    \label{t:gnn}
    \begin{tabular}{ c rrrr}
    \toprule
   \multirow{2}{*}{\# nodes.} & \multicolumn{2}{c}{Improv.} & \multicolumn{2}{c}{$\ell_2$ distance}\\
               & ours&GNN&ours&GNN\\
    \midrule
    $10^3$ & $\uparrow45.7\%$&$\uparrow 1.4\%$&0.05& 0.51\\
    $10^4$ & $\uparrow67.6\%$&$\uparrow 19.3\%$&0.2& 1.38\\
    $10^5$ & $\uparrow71.3\%$&\red{$\downarrow 4.0\%$}&0.95&30.35 \\
    \bottomrule
    \end{tabular}
    \label{portion_of_GPU}
    \end{table}
}

\section{Conclusions and Future Direction}
Inspired by recent advances in FOMs for LP problems and deep unrolling, 
this work introduces PDHG-Net, based on the PDHG algorithm, and proves that PDHG-Net can represent LP optimal solutions approximately. 
Building upon this architecture, we propose a
two-stage inference framework for solving large-scale LP problems. 
Comprehensive experiments demonstrate that our proposed method significantly improves the speed of solving large-scale LP problems when there are enough historical data. 
This indicates the potential of applying L2O to solving large-scale LP instances. 
Our result also sheds light on the architecture design
in L2O for other problems such as  mixed integer programming (MIP): while GNN has been the architecture of choice
in L2O for MIP, we propose a new architecture that has some similarities with GNN but enjoys a strong theoretical guarantee. 
A future direction is to explore our PDHG-Net in L2O for MIP. 

\section*{Acknowledgments}
Bingheng Li, Linxin Yang, Yupeng Chen, Senmiao Wang, Qian Chen, Akang Wang, Tian Ding, and Ruoyu Sun are supported by the National Key R\&D Program of China under grant 2022YFA1003900; 
Hetao Shenzhen-Hong Kong Science and Technology  Innovation Cooperation Zone Project (No.HZQSWS-KCCYB-2024016); 
University Development Fund UDF01001491, the Chinese University of Hong Kong, Shenzhen;
Guangdong Provincial Key Laboratory of Mathematical Foundations for Artificial Intelligence (2023B1212010001);
Longgang District Special Funds for Science and Technology Innovation (LGKCSDPT2023002).
\section*{Impact Statement}
This paper presents work whose goal is to advance the field
of Machine Learning. There are many potential societal
consequences of our work, none of which we feel must be
specifically highlighted here.
\normalem
\bibliography{example_paper}
\bibliographystyle{icml2024}

\newpage
\appendix
\onecolumn
\setcounter{equation}{0}
\renewcommand\theequation{\arabic{equation}}
\setcounter{equation}{0}
\section{Related Work}
\subsection{FOMs for Large-scale LP}
First-order methods (FOMs), which are based on gradient information rather than Hessian information, have been widely applied in numerous optimization fields and have become the standard approach~\cite{zhang2016first,kim2016optimized,beck2017first}. These methods are noted for their efficiency and simplicity in dealing with a variety of optimization problems, as evidenced by significant works such as
~\cite{renegar2014efficient,wang2017new,necoara2019linear}. In the realm of linear programming, the Simplex method ~\cite{dantzig1990origins} and Interior Point Methods (IPMs)~\cite{nesterov1994interior,ye2011interior,gondzio2012interior} have garnered attention for their exceptional performance in solving medium and small-sized LP problems. However, achieving additional scalability or acceleration remains a significant challenge. This challenge is particularly pronounced due to the necessity of solving linear systems, which leads to resource-intensive LU decomposition or Cholesky decomposition processes.

Recent research trends in the field of large-scale linear equation systems have shifted focus towards the exploration of FOMs. These methods are particularly attractive due to their low per-iteration cost and parallelization capabilities. Solvers based on FOMs have demonstrated notable success in empirical studies, as evidenced by the work of ~\cite{o2016conic, o2021operator,basu2020eclipse, lin2021admm, deng2022new}. This shift highlights the growing interest in and effectiveness of FOMs for handling large-scale computational challenges.

It is noteworthy that within the domain of FOMs, the Primal-Dual Hybrid Gradient (PDHG) method has exhibited faster convergence rates by effectively utilizing data matrices for matrix-vector multiplication. This advantage is well-documented in recent literature, including studies by ~\cite{applegate2023faster,hinder2023worst,lu2023geometry}. Specifically, PDHG demonstrates sublinear rates on general convex-concave problems, as shown in ~\cite{chambolle2016ergodic,lu2023unified}, and achieves linear rates in many instances ~\cite{fercoq2021quadratic,lu2022infimal}. From a practical application standpoint, the PDLP, a universal LP solver based on the PDHG algorithm, has achieved state-of-the-art performance in large-scale LP problems~\cite{applegate2021practical, lu2023cupdlp, cuPDLP-C}.
\subsection{Expressive Power of L2O architectures for LP problems}
With the advent of deep learning, learning to optimize (L2O) has become an emerging research area ~\cite{nair2020solving,bengio2021machine,chen2022learning}. Numerous studies have approached optimization problems as inputs, employing trained parametric models to provide valuable information for traditional optimizers. In the field of LP, these enhancements include advanced pre-processing techniques ~\cite{li2022learning, kuang2023accelerate}, 
and the generation of high-quality initial solutions~\cite{fan2023smart}. Additionally, these models have been utilized to yield some approximate optimal solutions for the LP problems themselves~\cite{chen2022representing,qian2023exploring}. Nevertheless, these methods are often tightly coupled with traditional algorithms, and their effectiveness is called into question with increasing problem sizes.

\section{Proof of Theorem \ref{alignment}}
\begin{proof}

We will prove a stronger claim by mathematical induction: there is a parameter assignment fulfilling the following requirement. For any hidden layer $k$, there exist (data-independent) matrices $P_x^k, P_y^k \in \mathbb{R}^{d_k \times d_k}$ satisfying the following property. Given any LP problem $\mathcal{M} = (G; l,u,c; h)$ and its corresponding primal-dual sequence $(x^k, y^k)_{k\leq K}$ generated by PDHG algorithm within $K$ iterations, the primal embedding $X^k$ and the dual embedding $Y^k$ at the $k$-th layer satisfy 
\begin{equation}\label{ind-hypo-rep}
\begin{aligned}
    X^k P_x^k &= [\bar{x}^k, x^k,l,u,c, \mathbf{0}_n, \dots \mathbf{0}_n] \in \mathbb{R}^{n \times d_k},\\
    Y^k P_y^k &= [\bar{y}^k, y^k,h,\mathbf{0}_n, \dots \mathbf{0}_n] \in \mathbb{R}^{n \times d_k},
\end{aligned}
\end{equation}
where
\begin{equation*}
    \bar{x}^k = \frac{1}{k} \sum_{i=1}^k x^i, \quad \bar{y}^k = \frac{1}{k} \sum_{i=1}^k y^i, \quad 1 \leq i \leq K.
\end{equation*}

Suppose that $G$ is an $m \times n$ real matrix. If this claim is true, $\bar{x}^k, x^k$ can be recovered by all the channels of $X^k$ through linear transformations
\begin{equation*}
\begin{aligned}
    \bar{x}^k &= (X^k P_x^k) \cdot e_1 = \underbrace{X^k}_{n \times d_k} \cdot (\underbrace{P_x^k e_1}_{d_k \times 1}), \\
    x^k &= (X^k P_x^k) \cdot e_2 = X^k \cdot (P_x^k e_2),
\end{aligned}
\end{equation*}
where $e_1 = (1,0,\dots, 0)^T$, $e_2 = (0,1,0,\dots, 0)^T$. Similarly, $\bar{y}^h, y^h$ can be recovered through $Y^h$ through
\begin{equation*}
\begin{aligned}
    \bar{y}^k &= (Y^k P_y^k) \cdot e_1 = \underbrace{Y^k}_{m \times d_k} \cdot (\underbrace{P_y^k e_1}_{d_k \times 1}), \\
    y^k &= (Y^h P_y^k) \cdot e_2 = Y^k \cdot (P_y^k e_2).
\end{aligned}
\end{equation*}
Moreover, the coefficient vectors of these linear channel combinations $P^k_x e_1, P^k_x e_2, P^k_y e_1, P^k_y e_2$ do not rely on $\mathcal{M}$ and $(x^k, y^k)_{k \leq K}$.

Now we prove this claim by mathematical induction.

1. Base case: $k = 0$. This claim is obviously true because the inputs are $X^0 = [x^0, l, u, c]$ and $Y^0 = [y^0, q]$. 


2. By the induction hypothesis, we suppose that there is an assignment of parameters before the $k$-th layer such that (\ref{ind-hypo-rep}) holds, then we consider the $(k+1)$-th layer. 

We assume here that $l$ and $u$ are vectors with only finite entries. In practice, the bounds for the (primal) variable $x$ are constrained by the precision limitations of the computing machinery.

First, we will show that $X^{k+1}$ satisfies (\ref{ind-hypo-rep}). We define
\begin{equation*}
    Q_1 := \begin{bmatrix}
        1 & -1 &  0 & 0 & 0 & 0 & 0 & 0 & 0 & 0\\
        0 & 0 & 1 & 1 & 0 & 0 & 0 & 0 & 0 & 0 \\
        0 & 0 & -1 & 0 & 1 & -1 & 0 & 0 & 0 & 0  \\
        0 & 0 & 0 & -1 & 0 & 0 & 1 & -1 & 0 & 0  \\
        \tau & \tau & 0 & 0 & \tau & \tau & \tau & \tau & \tau+1 & \tau - 1
    \end{bmatrix} \in \mathbb{R}^{5 \times 10},
\end{equation*}
\begin{equation*}
    Q_2 := \begin{bmatrix}
        0 &  0 & \hdots & 0 \\
        \vdots & \vdots & \ddots & \vdots \\
        0 & 0 & \hdots & 0 \\
        \tau & \tau & \hdots & \tau
    \end{bmatrix} \in \mathbb{R}^{5 \times (d_{k+1} - 10)}.
\end{equation*}
and
\begin{equation*}
    Q_3 := \begin{bmatrix}
        0 & 0 & 0 & 0 \\
        0 & 0 & 1 & 1 \\
        0 & 0 & 0 & 0 \\
        0 & 0 & 0 & 0
    \end{bmatrix} \in \mathbb{R}^{4 \times 4}.
\end{equation*}

Then we get
\begin{equation*}
\begin{aligned}
    &\quad [\bar{x}^k, x^k, l,u,c] \cdot Q_1 \\
    &= [\bar{x}^k+\tau c, -\bar{x}^k + \tau c, x^k - l, x^k - u, l+ \tau c, -l +\tau c, u+\tau c, -u + \tau c, (\tau +1)c, (\tau -1)c] \in \mathbb{R}^{n \times 10},
\end{aligned}
\end{equation*}
\begin{equation*}
    [\bar{x}^k, x^k, l,u,c] \cdot Q_2 = [\tau c, \tau c, \dots \tau c] \in \mathbb{R}^{n \times (d_{k+1} - 10)},
\end{equation*}
and
\begin{equation*}
    [\bar{y}^k, y^k, h, \mathbf{0}_n] \cdot Q_3 = [\mathbf{0}_n, \mathbf{0}_n, y^k, y^k].
\end{equation*}

If we define
\begin{equation}\label{weight_x assign equ}
    U^k_x := P^k_x \hat{U}^k_x, \quad U^k_y := P^k_y \hat{U}^k_y,
\end{equation}
where
\begin{equation*}
    \hat{U}^k_x = \left[\begin{array}{c|c } 
	Q_1 & Q_2 \\ 
	\hline 
	\mathbf{0}_{(d_{k} - 5) \times 10} & \mathbf{0}_{(d_{k} - 5) \times (d_{k+1} - 10)}
    \end{array}\right] \in \mathbb{R}^{d_k \times d_{k+1}},
\end{equation*}
and
\begin{equation*}
    \hat{U}^k_y = \left[\begin{array}{c|c } 
	Q_3 & \mathbf{0}_{4 \times (d_{k+1} - 4)} \\ 
	\hline 
	\mathbf{0}_{(d_{k} - 4) \times 4} & \mathbf{0}_{(d_{k} - 4) \times (d_{k+1} - 4)}
    \end{array}\right] \in \mathbb{R}^{d_k \times d_{k+1}}.
\end{equation*}

By basic linear algebra, we have
\begin{equation*}
\begin{aligned}
    X^k U^k_x = X^k P^k_x \hat{U}^k_x &= [\bar{x}^k + \tau c, -\bar{x}^k + \tau c, x^k - l, x^k - u, \\
    &\qquad \quad l + \tau c,-l + \tau c, u + \tau c,-u + \tau c, (\tau + 1)c,(\tau-1)c, \tau c, \dots, \tau c] \in \mathbb{R}^{d_k \times d_{k+1}},\\
    Y^k U^k_y = Y^k P^k_y \hat{U}^k_y &= [\mathbf{0}_n, \mathbf{0}_n, y^k, y^k, \mathbf{0}_n, \dots, \mathbf{0}_n] \in \mathbb{R}^{d_k \times d_{k+1}}.
\end{aligned}
\end{equation*}

We assign $\tau_k = \tau$ and denote $\tilde{x}^k = x^k - \tau (c - G^\top y^k)$, then we get
\begin{equation*}
\begin{aligned}
    &\qquad X^k U^k_x - \tau_k( c \cdot \mathbf{1}^{\top}_{d_{k+1}} - G^\top Y^k U^k_y) \\
    &= [\bar{x}^k + \tau c, -\bar{x}^k + \tau c, x^k - l, x^k - u, l + \tau c,-l + \tau c, u + \tau c,-u + \tau c, (\tau + 1)c,(\tau-1)c, \tau c, \dots, \tau c] \\
    &\qquad \qquad -[\tau c, \tau c, \tau c, \tau c, \dots, \tau c] + \tau G^{\top} \cdot [\mathbf{0}_n, \mathbf{0}_n, y^k, y^k, \mathbf{0}_n, \dots, \mathbf{0}_n] \\
    &= [\bar{x}^k, -\bar{x}^k, x^k - \tau c + \tau G^{\top} y^k - l, x^k - \tau c + \tau G^{\top} y^k - u, l, -l, u,-u, c,-c, \mathbf{0}_n, \dots, \mathbf{0}_n] \\
    &= [\bar{x}^k, -\bar{x}^k, \tilde{x}^k - l, \tilde{x}^k - u, l, -l, u, -u, c, -c, \mathbf{0}_n, \dots, \mathbf{0}_n],
\end{aligned}
\end{equation*}
and further
\begin{equation*}
\begin{aligned}
    X^{k+1} &= {\rm ReLU} \big(X^k U^k_x - \tau_k( c \cdot \mathbf{1}^{\top}_{d_{k+1}} - G^\top Y^k U^k_y) \big) \\
    &= [(\bar{x}^k)^+, (\bar{x}^k)^-, (\tilde{x}^k - l)^+, (\tilde{x}^k - u)^+, l^+, l^-, u^+, u^-, 
    c^+, c^-, \mathbf{0}_n, \dots, \mathbf{0}_n].
\end{aligned}
\end{equation*}

Since
\begin{equation}\label{01301604}
\begin{aligned}
    z &= z^+ - z^-, \\
    \mathrm{\bf{Proj}}_{l \leq z \leq u} (z) &= l + (z - l)^+ - (z-u)^+, \\
    \bar{x}^{k+1} &= \frac{k}{k+1} \bar{x}^k + \frac{1}{k+1} x^{k+1},
\end{aligned}
\end{equation}
for any $z \in \mathbb{R}^n$, we get
\begin{equation}\label{01301605}
\begin{aligned}
    x^{k+1} &= \mathrm{\bf{Proj}}_{l \leq z \leq u} (\tilde{x}^k) \\
    &= l + (\tilde{x}^k - l)^+ - (\tilde{x}^k-u)^+ \\
    &= l^+ - l^- + (\tilde{x}^k - l)^+ - (\tilde{x}^k-u)^+,
\end{aligned}
\end{equation}
and
\begin{equation}\label{01301606}
\begin{aligned}
    \bar{x}^{k+1} &= \frac{k}{k+1} \bar{x}^k + \frac{1}{k+1} x^{k+1} \\
    &= \frac{k}{k+1} (\bar{x}^k)^+ - \frac{k}{k+1} (\bar{x}^k)^- + \frac{1}{k+1} l^+ - \frac{1}{k+1} l^- + \frac{1}{k+1} (\tilde{x}^k - l)^+ - \frac{1}{k+1} (\tilde{x}^k-u)^+.
\end{aligned}
\end{equation}

We define
\begin{equation*}
    P^{k+1}_x = \left[\begin{array}{c|c } 
	Q_4 & \mathbf{0}_{10 \times (d_{k+1} - 5)} \\ 
	\hline 
	\mathbf{0}_{(d_{k+1} - 10) \times 5} & \mathbf{0}_{(d_{k+1} - 10) \times (d_{k+1} - 5)}
    \end{array}\right] \in \mathbb{R}^{d_{k+1} \times d_{k+1}},
\end{equation*}
where
\begin{equation*}
    Q_4 := \begin{bmatrix}
        \frac{k}{k+1} & 0 & 0 & 0 & 0 \\
        \frac{-k}{k+1} & 0 & 0 & 0 & 0 \\
        \frac{1}{k+1} & 1 & 0 & 0 & 0 \\
        \frac{-1}{k+1} & -1 & 0 & 0 & 0 \\
        \frac{1}{k+1} & 1 & 1 & 0 & 0 \\
        \frac{-1}{k+1} & -1 & -1 & 0 & 0 \\
        0 & 0 & 0 & 1 & 0 \\
        0 & 0 & 0 & -1 & 0 \\
        0 & 0 & 0 & 0 & 1 \\
        0 & 0 & 0 & 0 & -1
    \end{bmatrix} \in \mathbb{R}^{10 \times 5}.
\end{equation*}
Combining (\ref{01301604}), (\ref{01301605}), and (\ref{01301606}), we have
\begin{equation*}
\begin{aligned}
    X^{k+1} P^{k+1}_x 
    = [\bar{x}^{k+1}, x^{k+1}, l, u, c, \mathbf{0}_n, \dots, \mathbf{0}_n].
\end{aligned}
\end{equation*}
Obviously, $P^{k+1}_x$ is independent of the LP problem $\mathcal{M} = (G; l,u,c; h)$ and the primal-dual sequence $(x^k, y^k)_{k \leq K}$.

Second, we will show that $Y^{k+1}$ satisfies (\ref{ind-hypo-rep}). Similar to the induction of $X^{k+1}$, there exist data-independent matrices $\hat{V}^k_x, \hat{V}^k_y \in \mathbb{R}^{d_{k} \times d_{k+1}}$, and $\hat{W}^{k}_x \in \mathbb{R}^{d_{k+1} \times d_{k+1}}$ such that
\begin{equation*}
\begin{aligned}
    Y^k V^k_y &=  Y^k P^k_y \hat{V}^k_y = [\bar{y}^k - \sigma h, -\bar{y}^k - \sigma h, y^k, (1-\sigma)h, -(1+\sigma)h, - \sigma h, \dots, - \sigma h], \\
    X^{k+1} W^{k}_x &= X^{k+1} P^{k+1}_x \hat{W}^{k+1}_x = [\mathbf{0}_n, \mathbf{0}_n, x^{k+1}, \mathbf{0}_n, \dots, \mathbf{0}_n], \\
    X^{k} V^{k}_x &= X^{k} P^{k}_x \hat{V}^{k}_x = [\mathbf{0}_n, \mathbf{0}_n, x^{k}, \mathbf{0}_n, \dots, \mathbf{0}_n].
\end{aligned}
\end{equation*}
Here we assign the weight matrices
\begin{equation*}
     V^{k}_x = P^k_x \hat{V}^k_x, \qquad V^k = P^k_y \hat{V}^k_y, \qquad W^{k}_x = P^{h+1}_x \hat{W}^{k}_x. 
\end{equation*}


If we assign $\sigma_k = \sigma$, we get
\begin{equation*}
\begin{aligned}
     &\qquad Y^k V^k_y + \sigma_k (h \cdot \mathbf{1}^{\top}_{d_{k+1}} - 2G X^{k+1} W^{k}_x + G X^k V^k_x ) \\
    &= [\bar{y}^k - \sigma h, -\bar{y}^k - \sigma h, y^k, h- \sigma h, -h- \sigma h, - \sigma h, \dots, - \sigma h] \\
    & \qquad \qquad + [\sigma h, \sigma h, \sigma h, \dots, \sigma h] -\sigma G \cdot [\mathbf{0}_m, \mathbf{0}_m, 2x^{k+1} - x^k, \mathbf{0}_m, \dots, \mathbf{0}_m] \\
    &= [\bar{y}^k, -\bar{y}^k, y^k + \sigma (h - G(2x^{k+1} - x^k)), h, -h, \mathbf{0}_n, \dots, \mathbf{0}_n]
\end{aligned}
\end{equation*}
and further
\begin{equation*}
\begin{aligned}
Y^{k+1} &= {\rm ReLU}\big( Y^k V^k_y + \sigma_k (h \cdot \mathbf{1}_{d_{k+1}} - 2 G X^{k+1} W^{k}_x + G X^k V^k_x ) \big) \\
    &= [(\bar{y}^k)^+, (\bar{y}^k)^-, y^{k+1}, q^+, q^-, \mathbf{0}_n, \dots, \mathbf{0}_n].
\end{aligned}
\end{equation*}
Similar to the induction of $X^{k+1}$, there exists a data-independent $P^{k+1}_y$ such that
\begin{equation*}
\begin{aligned}
     Y^{k+1} P^{k+1}_y
    = [\bar{y}^{k+1}, y^{k+1}, h, \mathbf{0}_m, \dots, \mathbf{0}_m].
\end{aligned}
\end{equation*}

3. By the mathematical induction, we get that the claim (\ref{ind-hypo-rep}) holds for any hidden layer. In terms of the output layer $K$, we only need to replace the linear transformations after the ReLU activations $P^K_x$ and $P^K_y$ with $P^K_x e_1$ and $P^K_y e_1$, respectively. These small modifications ensure $X^K = \bar{x}^K$ and $Y^K = \bar{y}^K$.

\end{proof}

\section{Proof of Theorem \ref{poly expressive thm}}

\begin{proof}
    Given the approximation error bound $\epsilon$, we consider a $K(\epsilon)$-layer PDHG-Net with network width $d_k = 20 \geq 10$. We assign the parameter $\Theta_{\rm PDHG}$ as chosen in Theorem 1. By Theorem 1, for any LP problem $\mathcal{M} = (G; l, u,c;h)$ and its corresponding primal-dual sequence $(x^k, y^k)_{k \leq K(\epsilon)}$, the network output $(X^K, Y^K)$ is equal to $(\bar{x}^k, \bar{y}^k)$.

    Recalling Proposition \ref{pock_proposition}, for any $(x,y) \in \mathbb{R}^n \times \mathbb{R}^m_{\geq 0}$ satisfying $l \leq x \leq u$, we have
    \begin{equation*}\label{02011428}
    \begin{aligned}
        L(X^K, y; \mathcal{M}) - L(x, Y^K; \mathcal{M}) &= L(\bar{x}^k, y; \mathcal{M}) - L(x, \bar{y}^k; \mathcal{M}) \\
        &\leq \frac{C}{2K(\epsilon)},
    \end{aligned}
    \end{equation*}
    where $C := \frac{\|x-x^0\|^2}{\tau} + \frac{\|y-y^0\|^2}{\sigma} - (y - y^0)^\top G (x-x^0)$ for simplicity.

    We only need to take $K(\epsilon) = [C / \epsilon]+1$, then we get
    \begin{equation*}\label{02011428}
    \begin{aligned}
        L(X^K, y; \mathcal{M}) - L(x, Y^K; \mathcal{M}) \leq \epsilon / 2 < \epsilon.
    \end{aligned}
    \end{equation*}

    Since the network width is a fixed number $20$, the number of neurons at each layer is bounded by a constant. Thus, the total number of neurons has the same order as $K(\epsilon)$, which is $\mathcal{O}(1/\epsilon)$.

\end{proof}

\section{Feature extraction}\label{app:feat}
This section illustrates all used features to extract initial encoding of LP instances.
We used $3$-dimension feature vector for variables, $4$-dimensions for constraints and a single scalar value to describe the connection between variables and constraints.
Table~\ref{appt:feature} lists the detailed information of each utilized feature.

\begin{table}[htbp]
    \centering
    \begin{tabular}{c c c c c}
    \toprule
        & feature & type & dimension & description \\
        \midrule
        \multirow{3}{*}{Variables} & lb & float& 1 & the lower bound of the corresponding variable\\
         & ub & float& 1 & the upper bound of the corresponding variable\\
         & c & float& 1 & the cost coefficient of the corresponding variable\\
         \midrule
        \multirow{2}{*}{Constraints} & ctype & $\{0,1\}$ & 3 & One-hot encoding of the type of the constraint ([$\leq,=,\geq$]).\\
         & rhs & float & 1 & the right-hand-side value of the corresponding constraint\\
         \bottomrule
    \end{tabular}
    \caption{initial feature encoding of the LP instance.}
    \label{appt:feature}
\end{table}

\section{Datasets}\label{app:dataset}
\subsection{PageRank}
The generation of PageRank instances adheres to the methodology described in ~\cite{applegate2021practical}, wherein each problem is formulated into an LP  following the approach in~\cite{nesterov2014subgradient}. 
We employ the identical network configurations as in~\cite{applegate2021practical}, but vary the number of nodes to create instances of different sizes.
Sizes of all utilized instances can be found in Table~\ref{t:prsizes}.
\begin{table}[!htbp]
    \centering
    \begin{tabular}{c ccc}
    \toprule
    \# nodes & \# vars. & \# cons. & \# nnz.\\
    \midrule
    $10^3$&1,000&1,001&7,982\\
    $10^4$&10,000&10,001&79,982\\
    $5\times10^4$&50,000&50,001&399,982\\
    $10^5$&100,000&100,001&799,982\\
    $10^6$&1,000,000&1,000,001&7,999,982\\
    \bottomrule
    \end{tabular}
    \caption{Sizes of utilized PageRank instances.The number of variables, constraints, and non-zeros are reported.}
    \label{t:prsizes}
\end{table}
\subsection{Item Placement and Workload Appropriation}
We adopted the generation code from~\cite{gasse2022machine}, and modified controlling parameters to obtain instances with complex LP relaxations.
Specifically, in the \texttt{IP-S} dataset, each item is set to have 100 dimensions with a configuration of 50 bins and 525 items. 
Conversely, for the \texttt{IP-L} dataset, items are set to have 200 dimensions, with an increased scale of 150 bins and 1,575 items.
In the \texttt{WA-S} and \texttt{WA-L} datasets, we enhanced the probability of each task being compatible with corresponding machines, which leads to a much denser model, consequently increasing the complexity of the LP relaxation and making it more challenging to solve.
Furthermore, we adjust the quantities of machines and tasks to 800 and 100, and 2,000 and 220, respectively.
Sizes of all utilized instances can be found in Table~\ref{t:ml4cosize}.
\begin{table}[htbp]
    \centering
    \begin{tabular}{c ccc}
    \toprule
    dataset & \# vars. & \# cons. & \# nnz.\\
    \midrule
    \texttt{IP-S}& 31,350 & 15,525 & 5,291,250 \\
    \texttt{IP-L}& 266,450 & 91,575 & 94,826,250\\
    \texttt{WA-S}& 80,800 & 98,830 & 3,488,784 \\
    \texttt{WA-L}& 442,000 & 541,058 & 45,898,828\\
    \bottomrule
    \end{tabular}
    \caption{Sizes of utilized instances.The average number of variables, constraints, and non-zeros are reported.}
    \label{t:ml4cosize}
\end{table}
\section{Comparison with Gurobi}
Moreover, it's crucial to point out that Gurobi faced substantial difficulties in resolving instances with node counts exceeding $10^5$. 
This observation further highlights the superior capability of FOMs in efficiently tackling large-scale problems. 
The collected data from these experiments distinctly demonstrate that our proposed method consistently outperforms both PDLP and Gurobi.
Furthermore, we also conducted the comparitive experiments between our framework and the ABIP method, which is an Interior point method accelerated by FOM.

\setlength{\tabcolsep}{1.2mm}{
\begin{table*}[h]
\centering
\caption{Solve time for PageRank instances. Gurobi's LP solvers are set to default and use 1 thread. The precision of PDLP is set to the $10^{-8}$ relative gap. A time limit of 10 hours is set for all tested approaches, and presolve not applied. The number of nonzero coefficients per instance is $8 \times (\text{\# nodes}) - 18$.}
\label{t:pagerankgutobi}
\begin{tabular}{ccccccc c}
\toprule
       \# nodes &ours&PDLP &    Gurobi Barrier  & Gurobi P. Simp. &Gurobi D. Simp.  & ABIP \\
\midrule
$10^3$  &0.01s &0.04s&       0.04s & 0.1s & 0.2s & 0.05s\\
$10^4$  &0.4s &1.1s&                             273.9s & 40.9s & 125.8 s  & 3.73s\\
$10^5$  &22.4s &71.3s&                          \textgreater 15,000s & \textgreater 15,000s & \textgreater 15,000s & 277.3s\\
$10^6$  &4,508s&\textgreater 15,000s &                       \textgreater 15,000s &  \textgreater 15,000s & \textgreater 15,000s & Fail\\
\bottomrule
\end{tabular}
\end{table*}
}

\end{document}